\documentclass[twocolumn]{article}

\usepackage[utf8]{inputenc}
\usepackage{amsmath}
\usepackage{graphicx}
\usepackage[a4paper, total={6.5in, 9in}]{geometry} 
\usepackage{authblk} 
\usepackage{hyperref} 
\usepackage{subcaption} 
\usepackage{url}
\renewenvironment{abstract}
  {\begin{center}\textbf{Abstract}\end{center}\itshape}
  {}
\usepackage{titlesec}
\titleformat{\section}[block]{\centering\bfseries\large}{\thesection}{1em}{}
\titleformat{\subsection}[block]{\bfseries\normalsize}{\thesubsection}{1em}{}

\title{\textbf{Car Object Counting and Position Estimation via Extension of the CLIP-EBC Framework}}

\author[1]{Seoik Jung}
\author[2]{Taekyung Song}

\affil[1]{Department of Software Engineering, Chungbuk National University}
\affil[2]{Department of Artificial Intelligence, Chung-Ang University \vspace{1em}} 
\affil[ ]{\texttt{acc6452@gmail.com}, \texttt{songteagyong@gmail.com}}

\date{} 

\begin{document}

\maketitle

\begin{abstract}
\noindent In this paper, we investigate the applicability of the CLIP-EBC framework, originally designed for crowd counting, to car object counting using the CARPK dataset. Experimental results show that our model achieves second-best performance compared to existing methods. In addition, we propose a K-means weighted clustering method to estimate object positions based on predicted density maps, indicating the framework's potential extension to localization tasks.
\end{abstract}

\section{Introduction}

Vision-language models like CLIP have demonstrated excellent performance in various recognition tasks based on their powerful ability to connect images and text [1]. CLIP-EBC is a framework proposed to solve crowd density estimation and counting problems by leveraging CLIP's text encoder [2]. Traditional classification-based counting methods struggled with continuous value labels and suffered from label boundary ambiguity. CLIP-EBC addresses these issues by introducing an integer-based quantization strategy and a density map-based loss, significantly improving the accuracy of crowd density estimation. Notably, CLIP-EBC achieved high performance and versatility by training only the image encoder while keeping the pre-trained text encoder of the CLIP model frozen.

This study aims to extend the applicability of the CLIP-EBC framework beyond crowd counting to a non-crowd domain: car object counting. Car object counting has high utility in various fields such as parking space management and traffic flow analysis, and the object distribution characteristics differ from those of crowds. To this end, we applied the CLIP-EBC model to the representative car object counting dataset, CARPK, to evaluate its counting performance. Through this, we confirmed that CLIP-EBC, originally optimized for crowd counting, can achieve meaningful results in a non-crowd domain.

Furthermore, beyond predicting simple block-level counts, we also explored the potential for extending the framework to estimate the central position (localization) of individual objects using the predicted count values. For this purpose, we apply a K-means-based clustering technique [5] to effectively extract the positions of car objects from the density map. This approach demonstrates the potential to expand the utility of vision-language-based counting models to various object counting and position estimation tasks beyond crowds.

\section{Methodology}

\subsection{Related Work}

Previous research in object counting has primarily evolved in two directions. One is the detection-based method, which detects objects first and then counts them, such as YOLO [9] and Faster R-CNN [10]. This method has limitations, as its performance degrades in dense scenes with severe object overlap. The other is the regression-based method, which predicts the total number of objects by converting an image into a density map.

Recently, research aiming for class-agnostic object counting has been actively pursued. For example, CounTR [3] utilizes a Transformer-based architecture [11] to achieve generalized counting performance across various object categories. Additionally, SAFECount [4] proposed a method for estimating the number of objects through similarity-based feature enhancement, even in few-shot scenarios. However, these methods are not optimized for environments with relatively low density and structured object distributions, such as cars.

Meanwhile, car object counting is typically performed on a specific object (cars) in a constrained environment like a parking lot, with the CARPK dataset [8] being a prime example. Most existing studies have relied on detection-based or regression-based counting and did not consider counting using a text encoder or performing additional position estimation.

\subsection{Problem Definition}

The existing CLIP-EBC framework was developed with a focus on performing crowd density estimation and counting by leveraging a pre-trained CLIP text encoder and image encoder. While this enabled effective crowd count prediction based on the powerful recognition capabilities of the CLIP model, it has the following limitations.

First, the model is optimized for a specific domain—crowd counting—and its potential for extension to other types of objects or environments has not been sufficiently verified. Second, CLIP-EBC has a structural limitation in that it focuses on predicting the number of objects per block, thereby discarding individual object location information. This leads to a problem where the detailed spatial information of objects is not fully utilized in the density map prediction process.

\subsection{Proposed Method}

Based on this problem awareness, we propose an approach that extends the CLIP encoder-based architecture from crowd counting to the car object counting problem and additionally incorporates an object position estimation function. To achieve this, we adapt the text prompts originally for crowd counting to sentences suitable for car object counting. We then induce domain adaptation by applying Visual-Prompt-Tuning (VPT) [6] to the image encoder while keeping the text encoder frozen. Furthermore, we estimate the central coordinates of individual objects by applying Gaussian filtering and K-means clustering [5] to the predicted density map. Through this, we aim to explore the possibility for a vision-language model to perform both counting and position estimation in various object environments.

\section{Implementation and Experiments}

\subsection{Model Architecture}

\begin{figure}[h!]
    \centering
    \includegraphics[width=\linewidth]{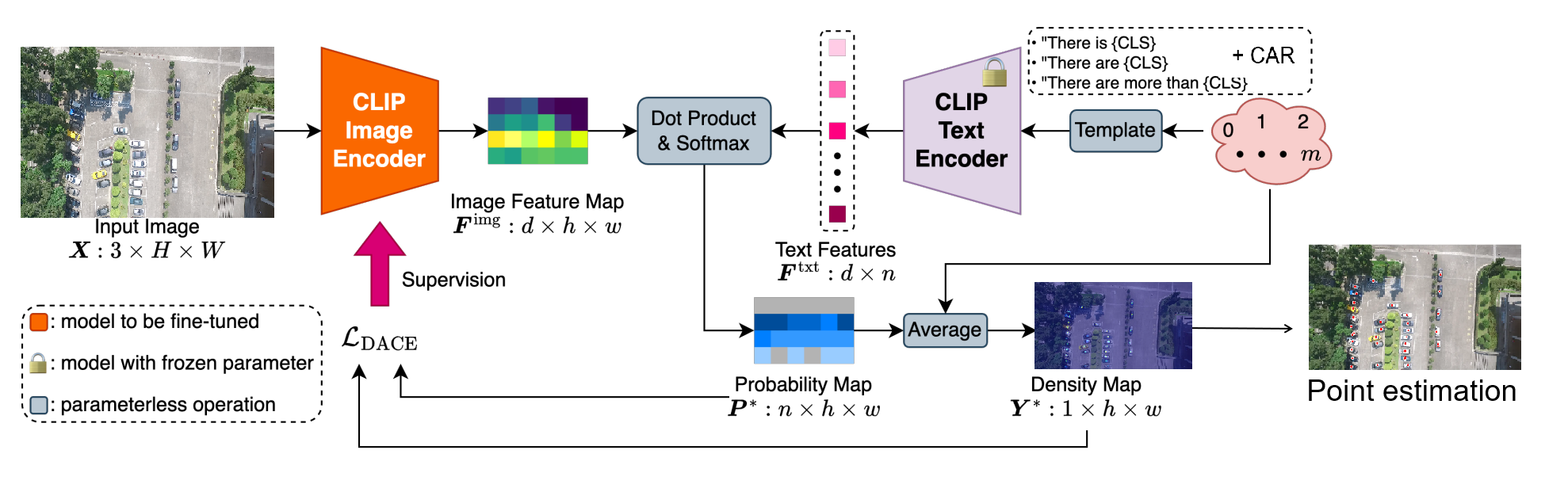} 
    \caption{The proposed architecture.}
    \label{fig:architecture}
\end{figure}

CLIP-EBC divides an input image into blocks and quantifies the number of objects within each block into integer bins, predicting a probability distribution based on similarity with text prompts. The resulting probability map is then converted into a final density map through an averaging operation. In this study, we applied a technique of inserting 32 Visual Prompt Tokens into each layer while keeping the CLIP's ViT-based [7] image encoder frozen. The text encoder maintains the pre-trained structure of CLIP, and the prompts are structured in the format "There are N cars". This approach induced effective domain adaptation without tuning the entire set of parameters. The overall architecture of the proposed CLIP-EBC-based model is illustrated in Figure \ref{fig:architecture}.

\subsection{Dataset}

To verify the applicability to the car object counting domain, this study used the Car Parking Lot (CARPK) dataset [8]. CARPK includes a total of 1,448 aerial images and approximately 90,000 car object labels. It features diverse object sizes and arrangements due to variations in parking lot structures and densities.

\subsection{Training and Experimental Setup}

Input images were resized to $224 \times 224$, and the block size was fixed at 8. The model was trained using the Adam optimizer with an initial learning rate of 1e-4 and a weight decay of 1e-4. The Distance-Aware Cross Entropy (LDACE) was used as the loss function. Training was conducted for a total of 2600 epochs with a batch size of 128. The training was performed on a single NVIDIA A100 GPU and took approximately 7 days. Additionally, a warm-up learning rate strategy was used, linearly increasing the learning rate for the first 50 epochs, followed by a cosine annealing scheduler to gradually decrease it.

Model performance was evaluated using Mean Absolute Error (MAE) and Root Mean Squared Error (RMSE). The performance comparison on the CARPK dataset is shown below. As seen in Table \ref{tab:performance}, the proposed CLIP-EBC model recorded the second-best MAE and RMSE, following HLCNN.

\begin{table}[h!]
    \centering
    \caption{Performance comparison of models on the CARPK dataset.}
    \label{tab:performance}
    \begin{tabular}{|l|c|c|}
        \hline
        \textbf{Model} & \textbf{MAE} & \textbf{RMSE} \\
        \hline
        HLCNN & 2.12 & 3.02 \\
        SAFECount & 5.33 & 7.04 \\
        CounTR & 5.75 & 7.45 \\
        BMNet+ & 5.76 & 7.83 \\
        VLCounter & 6.46 & 8.68 \\
        \textbf{CLIP-EBC-CAR (our)} & \textbf{4.01} & \textbf{6.02} \\
        \hline
    \end{tabular}
\end{table}

\subsection{Position Estimation}

\begin{figure}[h!]
    \centering
    \begin{subfigure}[t]{0.31\linewidth}
        \includegraphics[width=\linewidth]{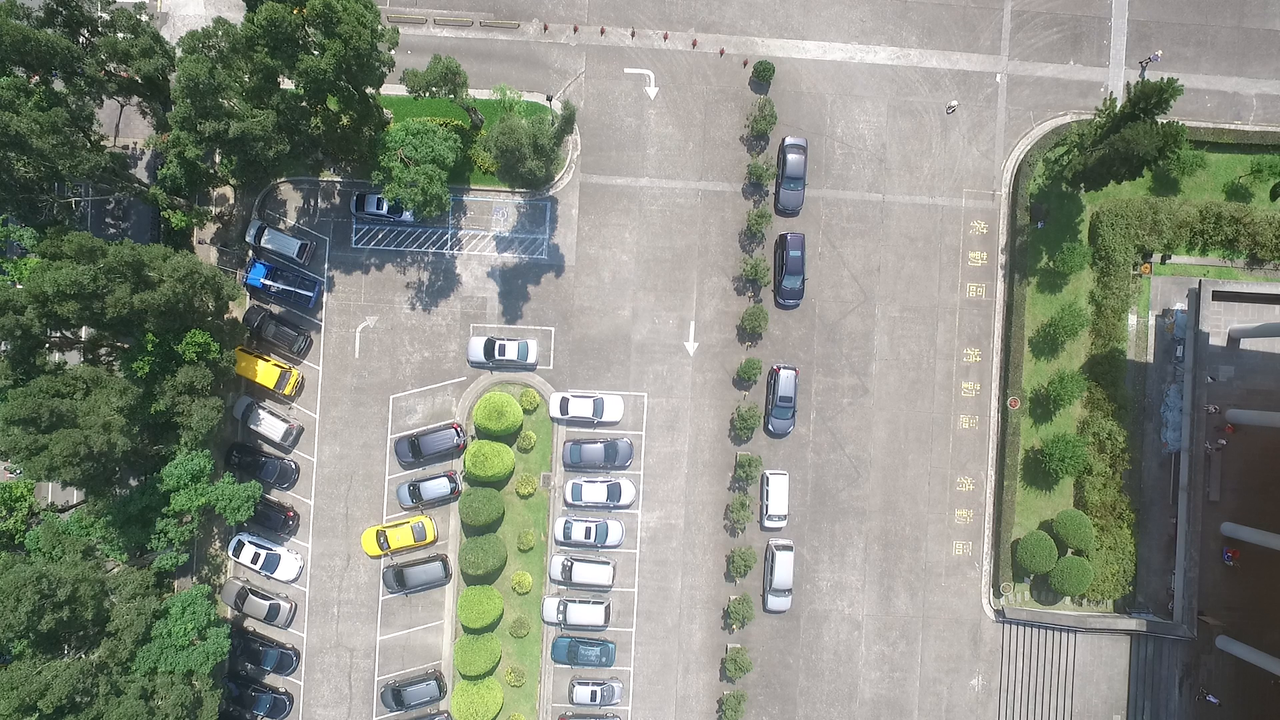}
        \caption{Original Image}
        \label{fig:2a}
    \end{subfigure}
    \hfill
    \begin{subfigure}[t]{0.31\linewidth}
        \includegraphics[width=\linewidth]{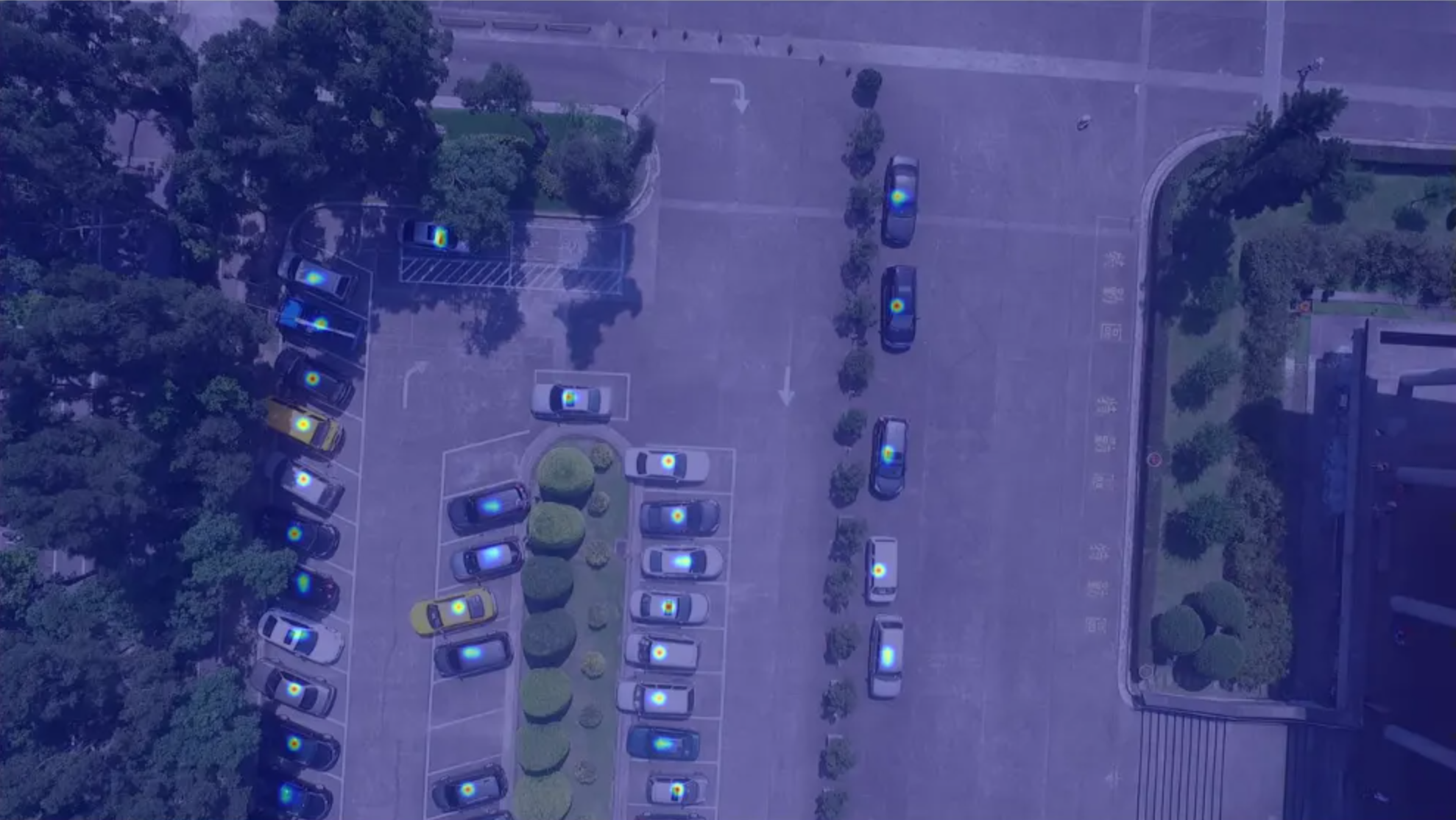}
        \caption{Density Map}
        \label{fig:2b}
    \end{subfigure}
    \hfill
    \begin{subfigure}[t]{0.31\linewidth}
        \includegraphics[width=\linewidth]{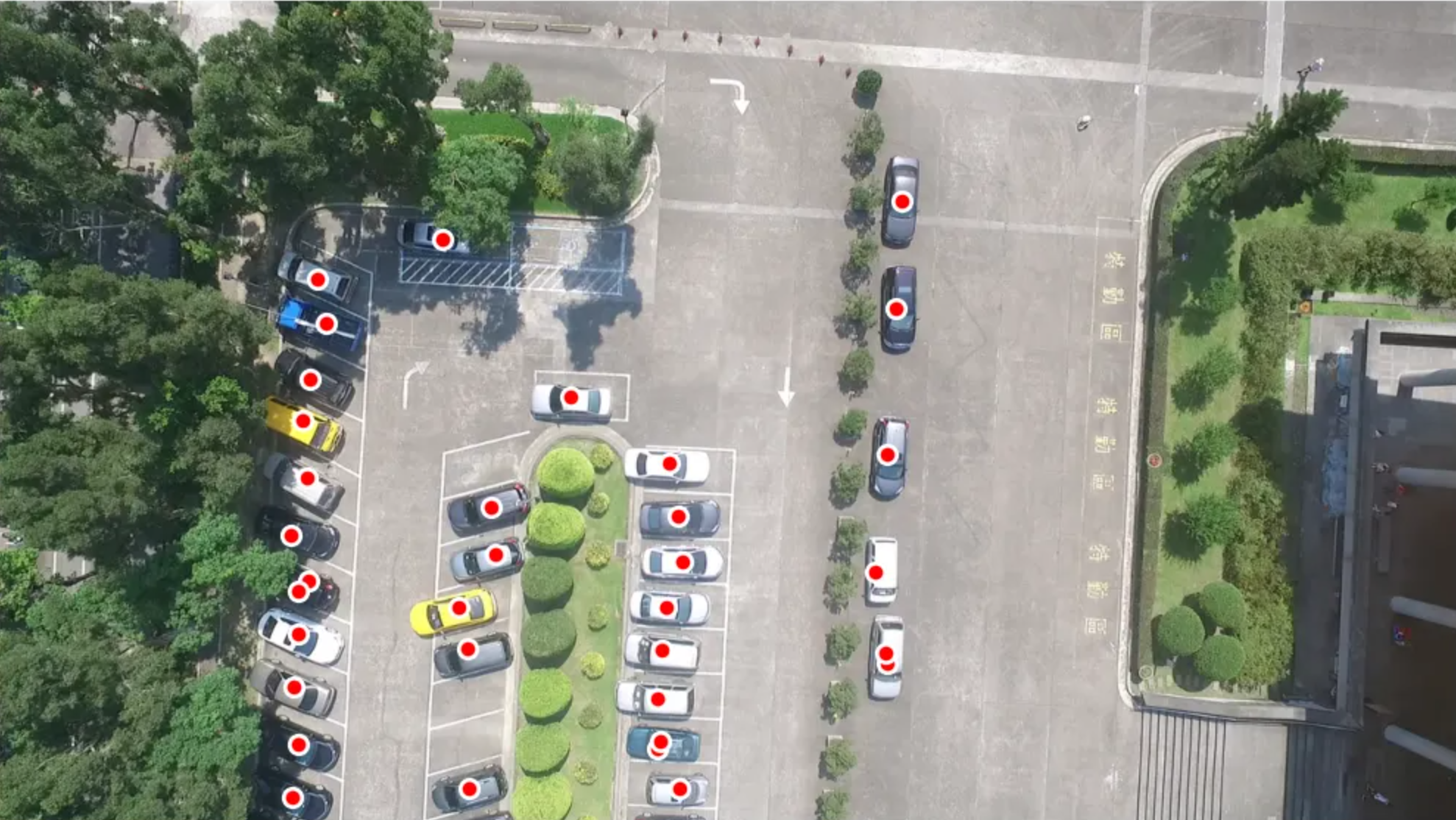}
        \caption{K-means}
        \label{fig:2c}
    \end{subfigure}
    \caption{Visualization results of position estimation on the CARPK dataset.}
    \label{fig:visualization}
\end{figure}

In this study, we extended the density map-based counting results by proposing a K-means-based algorithm to estimate the central position of individual objects. Specifically, a Gaussian filter is first applied to the predicted density map for smoothing, after which candidate pixels are selected based on an upper percentile threshold (e.g., top 3\%). Subsequently, the density values are normalized into probabilities, and candidate pixels are sampled in proportion to the predicted number of objects. Then, K-means clustering is performed, with the number of clusters set to the predicted object count. Finally, the center of each cluster is considered the object's position. This method complements the density-based counting results and shows the potential to restore the detailed spatial distribution of individual objects. Visualization results can be seen in Figure \ref{fig:visualization}.

\section{Conclusion}

In this study, we extended the applicability of the CLIP-EBC framework beyond the field of crowd counting to the non-crowd domain of car object counting. While maintaining the existing architecture and keeping the entire model parameters frozen, we attempted domain adaptation by only applying the VPT technique and modifying text prompts, achieving the second-best performance compared to the existing SOTA. This suggests that a model optimized for the crowd counting domain can be generalized to non-crowd domains through adaptation. Furthermore, we also presented the potential for object center position estimation by applying K-means clustering to the prediction results.


\end{document}